\documentclass[runningheads,a4paper]{llncs}

\usepackage{amssymb}
\usepackage[export]{adjustbox}
\usepackage{amsmath}
\usepackage{graphicx}
\usepackage{bbding}
\usepackage[multi-part-units=single]{siunitx}
\usepackage{url}
\usepackage{multirow}
\usepackage[misc]{ifsym}

\newcommand{\keywords}[1]{\par\addvspace\baselineskip
\noindent\keywordname\enspace\ignorespaces#1}

\begin{document}

\mainmatter  

\title{Needle Tip Force Estimation using an OCT Fiber and a Fused convGRU-CNN Architecture}

\titlerunning{Needle Tip Force Estimation with a convGRU-CNN Model}

%
%

\author{Nils Gessert\inst{1}$^\textrm{\Letter}$ \and Torben Priegnitz\inst{1} \and Thore Saathoff\inst{1}\and Sven-Thomas Antoni\inst{1} \and David Meyer\inst{2}\and Moritz Franz Hamann\inst{2}\and Klaus-Peter J\"unemann\inst{2} \and Christoph Otte\inst{1} \and Alexander Schlaefer\inst{1}}

\authorrunning{Nils Gessert et al.} 



\institute{$^1$Inst. of Medical Technology, Hamburg University of Technology, Hamburg, Germany\\
\email{nils.gessert@tuhh.de}\\
$^2$Department of Urology, University Hospital Schleswig-Holstein, Kiel, Germany}

%
%

\maketitle

\begin{abstract} 

Needle insertion is common during minimally invasive interventions such as biopsy or brachytherapy. During soft tissue needle insertion, forces acting at the needle tip cause tissue deformation and needle deflection. Accurate needle tip force measurement provides information on needle-tissue interaction and helps detecting and compensating potential misplacement. For this purpose we introduce an image-based needle tip force estimation method using an optical fiber imaging the deformation of an epoxy layer below the needle tip over time. For calibration and force estimation, we introduce a novel deep learning-based fused convolutional GRU-CNN model which effectively exploits the spatio-temporal data structure. The needle is easy to manufacture and our model achieves a mean absolute error of $\SI{1.76 \pm 1.5}{\milli\newton}$ with a cross-correlation coefficient of $0.9996$, clearly outperforming other methods. We test needles with different materials to demonstrate that the approach can be adapted for different sensitivities and force ranges. Furthermore, we validate our approach in an ex-vivo prostate needle insertion scenario. 

\keywords{Force Estimation, Optical Coherence Tomography, Convolutional GRU, Convolution Neural Network, Needle Placement}
\end{abstract}

\section{Introduction}

Needle insertion is widely used in minimally invasive procedures, e.g., for biopsies or brachytherapy. Automated needle insertion is challenging and includes aspects like image guidance, needle steering, and force measurement \cite{taylor2016medical}. Precise estimation of the forces acting on the needle tip is particularly interesting, e.g., for monitoring the needle-tissue interaction and detecting tissue ruptures, or to generate feedback during an intervention \cite{okamura.2004force}. One approach is to measure the forces at the needle shaft. While this allows for simple integration of conventional force-torque sensors, the measurements do not reflect the actual forces acting on the needle tip. As large frictional forces act on the needle shaft during insertion, force sensors need to be decoupled or located close to the tip in order to obtain accurate needle tip force estimates \cite{kataoka.2002measurement}. The small diameter of the needles complicates the integration of sensors \cite{rodrigues.2014influence}, which is particularly challenging for conventional mechatronic force sensors \cite{kataoka.2002measurement,Hatzfeld.2017}. In contrast, fiber optical force sensors are small, largely biocompatible, and not affected by electromagnetic interference, i.e., they are MRI-compatible \cite{beekmans.2016fiber}. Therefore, sensors based on Fabry-P\'erot interferometry \cite{beekmans.2016fiber} or Fiber Bragg Gratings \cite{kumar.2016detecting} have been proposed. Although these approaches have shown promising calibration results, they are rarely validated with tissue experiments \cite{mo.2017capability} and manufacturing and signal processing can be difficult, e.g., when the fibers are subjected to varying temperatures or lateral forces. Yet another approach is based on optical coherence tomography (OCT), where A-scan images of a cylindric instrument tip have been used to estimate the deformation and hence the strain acting on a translucent silicone layer \cite{kennedy2015quantitative}. \par

We consider force sensing using OCT and a sharp needle with a cone tip mounted on a needle using epoxy resin. The axial force acting on the tip is inferred from the epoxy layer's deformation observed in a series of A-scans. We can tailor our method to specific application scenarios by using softer epoxy resin for higher sensitivity, as required for microsurgery, or stiffer epoxy resin for larger forces, e.g., as occurring during biopsies \cite{beekmans.2016fiber}.
Generally, our approach is flexible and easy to manufacture as the epoxy material is interchangeable, the cone shape can be varied, and no accurate fiber placement is required. However, this imposes some challenges for calibration and force estimation, namely the robust identification of the deformation of the epoxy layer and a non-linear mapping of the measured deformations to forces. To this end we propose a force estimation method based on a novel convolutional gated recurrent unit-convolutional neural network (convGRU-CNN) architecture. Considering the high temporal sampling rate of OCT, we use a sequence of subsequent A-scans as an input to our model. In this way, we can take advantage of a rich spatio-temporal signal space for precise force estimates.

First, we present a detailed description of the force sensing needle and our convGRU-CNN architecture. Second, we describe our setups for calibration and evaluation of the needles. Third, we study the repeatability of force estimation for three different needles with different epoxy resin types. Finally, we present results for needle insertion into actual ex-vivo prostate tissue illustrating the feasibility of the approach and the importance of measurements at the needle tip.

\begin{figure}[t]
	\centering
	\includegraphics[width=1.0\columnwidth]{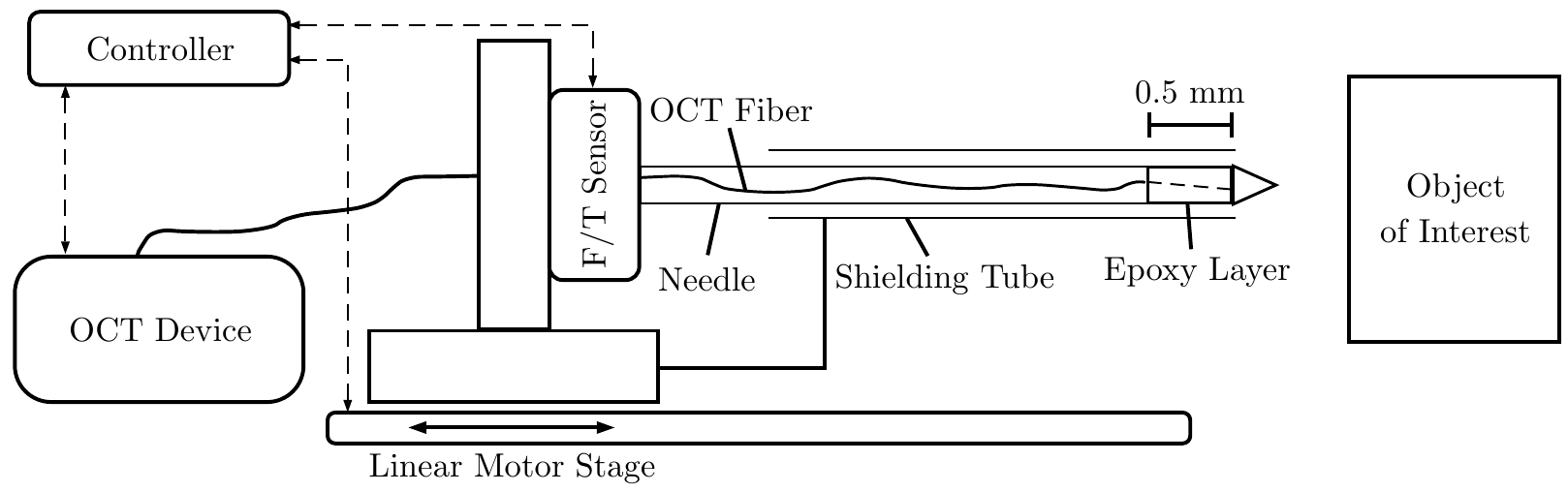}
	\caption{Schematic drawing of the needle and the experimental setup. Not to scale. The needle contains an OCT fiber that images a deformable epoxy layer at the needle tip. Forces are measured by the force sensor at the base. A shielding tube is decoupled from the needle and the force sensor and prevents shaft friction measurements for tissue insertion experiments. The setup is moved with a linear stage.}
	\label{fig:setup}
\end{figure}

\section{Materials and Methods}

\subsection{Needle Design and Experimental Setup}

\begin{figure}[t]
	\centering
	\adjincludegraphics[width=1.0\columnwidth,trim={0 {.05\height} 0 {.12\height}},clip]{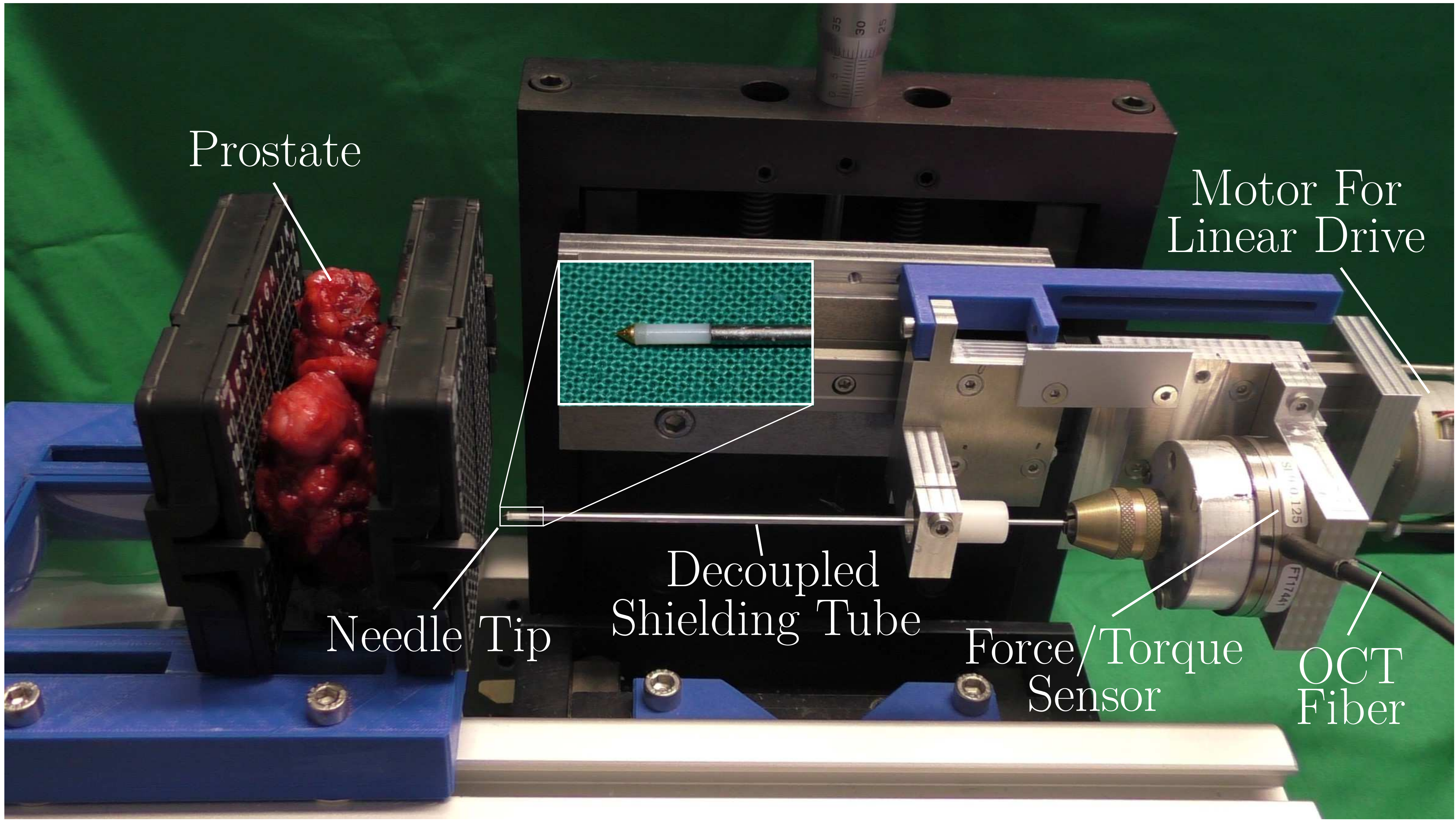}
	\caption{Photograph of the experimental setup for the prostate insertion experiment. Note, that for calibration the needle is driven against a metal plate.}
	\label{fig:setup_pic}
\end{figure}

A schematic drawing of our needle and the needle driver is shown in Figure~\ref{fig:setup}. The needle has a diameter of \SI{1.25}{\milli\metre}. An epoxy resin layer of approximately \SI{0.5}{\milli\metre} connects the cone shaped tip to the needle shaft. An optical fiber is embedded into the shaft and glued to the epoxy. The fiber is connected to an OCT device (Thorlabs Telesto I). A linear motion stage is used to drive the needle along its axial direction. For calibration and evaluation, a force sensor (ATI Nano43) is mounted between the needle shaft and the motion stage. 
To study how the sensitivity of the sensor can be varied by using different epoxy resin, the resin was mixed with Norland Optical Adhesive (NOA) 1625 in different concentrations. The layer and the needle tip on top are attached using NOA 63.
For calibration, the needle was driven against a metal plate. A large set of data was acquired by deforming the tip with random magnitude and velocity. 
For evaluation, the needle was inserted into a freshly resected human prostate at constant velocity. As the force sensor at the base measures the total force including friction, we consider a second setup using a shielding tube decoupled from the needle and the force sensor. In this way, we can measure the actual axial tip forces. Note, that this would be impractical for actual application as the tube is not flexible and would increase trauma. A photograph of the experimental setup is shown in Figure~\ref{fig:setup_pic}. We provide a video and ultrasound recording of two insertion procedures with pork liver and a human ex-vivo prostate in our supplementary material.



\subsection{Model Architecture} \label{sec:model}

\begin{figure}[t]
	\centering
	\includegraphics[angle=90,width=1.0\columnwidth]{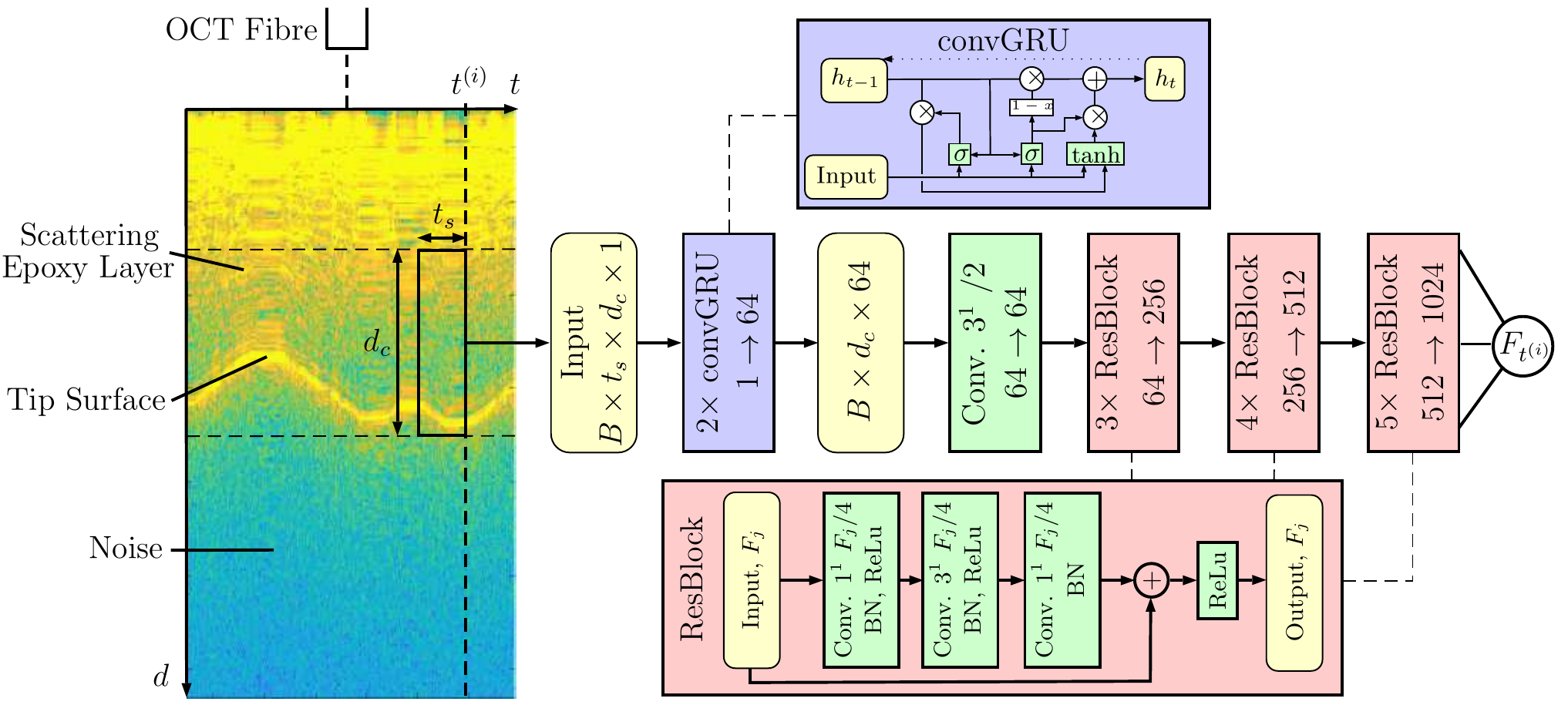}
	\caption{The convGRU-CNN model we employ. A series $t_s$ of cropped A-scans of size $d_c$ is fed into the model. The metal tip's lower surface cannot be penetrated by infrared light which is why this signal part is considered noise. The first block in a series of ResBlocks uses a stride of $2$ for the convolutions with kernel $3^1$ and increases the number of feature maps. Subsequent blocks have a stride of $1$ and keep the same feature map amount. The change in the number of feature maps is denoted in each group of ResBlocks. $F_j$ denotes the number of feature maps of ResBlock $j$. 
	}
	\label{fig:model}
\end{figure}

For our model input, we consider a series of A-scans prior to the current observation, as the current force estimate likely depends on prior deformation \cite{Aviles.2017towards}. Furthermore, we do not extract the epoxy surface as an explicit deformation feature but instead let our model learn relevant features. In this way, we avoid inconsistencies when extracting features for different materials and tips and we exploit information captured in the deformed epoxy layer itself.
 Prior approaches for spatio-temporal data used CNNs to extract features from image data which are fed into a recurrent model \cite{donahue.2015long}. Alternatively, the temporal dimension can also be handled by a convolution operation \cite{sun2015human}. Also, convolutional long-short term memory (convLSTM) cells have been introduced which allow for temporal processing of high dimensional structured data \cite{xingjian2015convolutional}. Based on these approaches, we propose a novel convGRU-CNN architecture, as shown in Figure~\ref{fig:model}. First, convGRU units take care of the temporal processing which results in a set of 1D feature maps. Then, a ResNet inspired \cite{He.2016} 1D CNN takes care of spatial processing.
Compared to LSTMs, GRUs merge the input and forget gate and they merge the cell and hidden state for higher efficiency. We use a combination of convLSTMs and GRUs by replacing the matrix multiplications with convolutions. 
The proposed architecture is compared to other approaches that have been introduced for spatio-temporal data processing. We consider a 2D CNN that processes both the temporal and the spatial dimension with convolutions. Its structure is the same as the CNN part in the convGRU-CNN model. Moreover, we consider a CNN-GRU model where the 1D CNN first processes the A-scans at each time step separately. Then, the CNN feature vector is fed into two standard GRUs. Next, we consider a pure 1D CNN that does not consider prior A-scans for the current force prediction. Last, we consider a pure 3-layer GRU model.
All networks are trained end-to-end in a single optimization run. We use the Adam algorithm for mini-batch gradient descent with a batch size of $B=100$. We implement our models using the Tensorflow environment. 

\subsection{Data Acquisition and Datasets}

OCT is an interferometry-based image modality using near infrared light to create 1D depth scans (A-scans) of up to $\SI{3}{\milli\metre}$ reflecting the inner structure of materials. We acquired A-scans at a rate of $\SI{5500}{\hertz}$ and force measurements at $\SI{500}{\hertz}$. We match the two data streams with nearest-neighbor interpolation, using the streams' timestamps. Our dataset consists of sequences of $t_s$ subsequent A-scans, each labled with a force measurement. Given that we do not need the full imaging depth of the OCT, image data beyond the maximum depth of the cone tip surface is cropped.

We consider calibration datasets for three needles with different epoxy resin types for evaluation with our convGRU-CNN model. Each dataset contains approximately $90000$ sequences of A-scans, each labeled with an axial force. By default we use a window of $t_s = 50$ with a crop size $d_c = 70$ pixels. We use $\SI{80}{\percent}$ of the data for training and validation sets, which we use to optimize hyperparamters, e.g., $t_s$, $d_c$, $l_r$, and network depth. The remaining $\SI{20}{\percent}$ of the data are used for testing. Sequences from the three sets are non-overlapping. 
Furthermore, one of the needles was evaluated in an ex-vivo experiment in a human prostate. We evaluate our proposed architecture by comparing it to the models mentioned in Section~\ref{sec:model}, reporting the mean absolute error (MAE), relative MAE (rMAE) and correlation coefficient (CC) between predictions and targets. 
All errors are given for the test set.


\section{Results}

\begin{figure}[t]
	\centering
	\adjincludegraphics[width=1.0\columnwidth,trim={0 0 0 {.008\height}},clip]{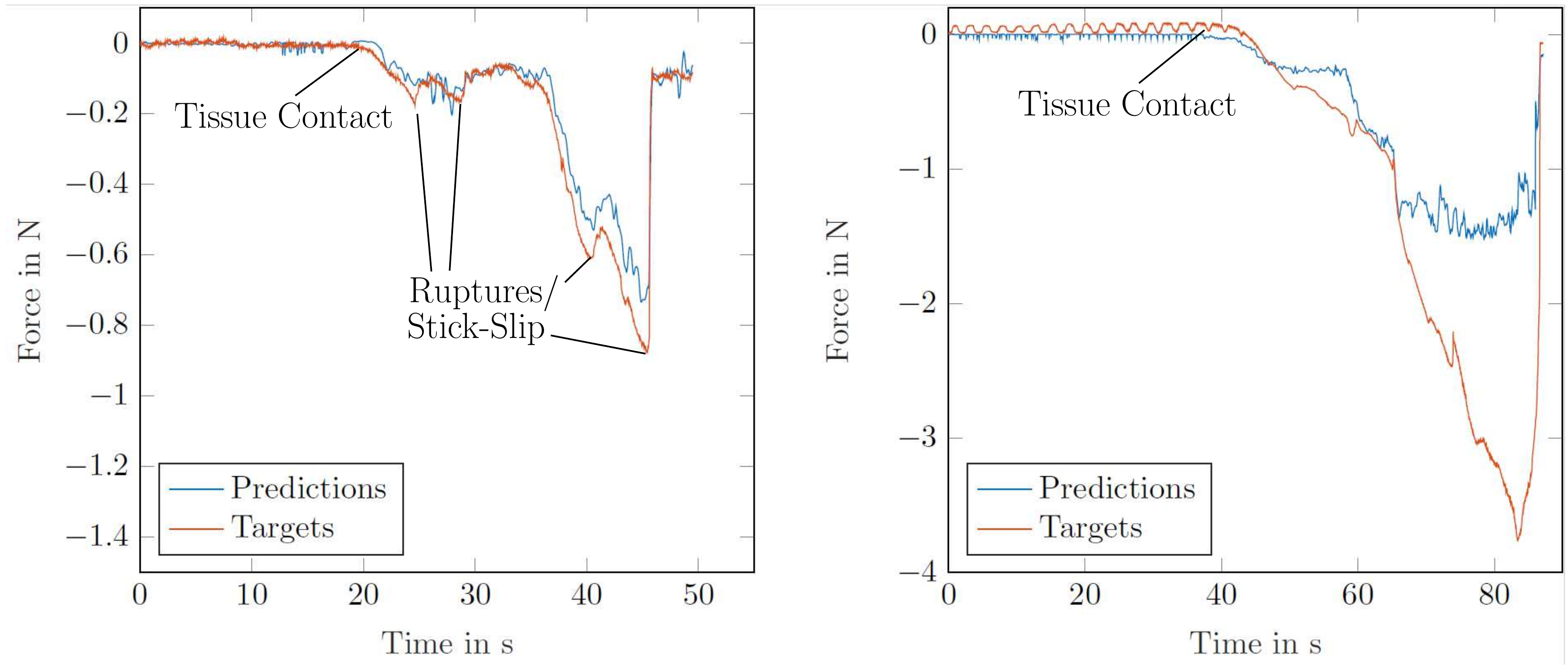}
	\caption{Predicted and measured force values are shown for an insertion with the shielding tube (left) and without (right). Note, that differences between predictions and targets are caused by friction during target measurements, not inaccurate model calibration. Needle 2 and the model convGRU-CNN are used for this experiment.}
	\label{fig:tissue}
\end{figure}

First, we report the results for different needles with a different stiffness of the epoxy layer. The results with the corresponding force magnitudes are shown in Table~\ref{tab:res_calibs}. The absolute error values vary, as the corresponding force ranges differ, however, the relative measures rMAE and CC show that models perform overall similar. 
\begin{table}
	\centering
	{\setlength{\tabcolsep}{1.5em}
	\begin{tabular}{l l l l l}
	 & MAE & rMAE & CC & Max \\ \hline
	Needle 1 & $1.76 \pm 1.5$ & $0.0213 \pm 0.0180$ & $0.9996$ & $379$ \\
	Needle 2 & $7.46 \pm 6.2$ & $0.0275 \pm 0.0231$ & $0.9994$ & $974$ \\	
	Needle 3 & $24.26 \pm 22.4$ & $0.0369 \pm 0.0322$ & $0.9989$ & $3202$ \\
 \hline \\
	\end{tabular}}
	\caption{Comparison of needles with different epoxy layer stiffnesses. The MAE  in $\si{\milli\newton}$ and rMAE (with standard deviation), the CC and the maximum force range in $\si{\milli\newton}$ are shown. The convGRU-CNN model was used for this experiment.}
	\label{tab:res_calibs}
\end{table}
Next, we present results for alternative model architectures. The results are shown in Table~\ref{tab:models}. The results show, that models that take prior A-scans into account perform better. Moreover, our proposed model outperforms previously introduced approaches for our application. 
\begin{table}
	\centering
	{\setlength{\tabcolsep}{1.5em}
	\begin{tabular}{l l l l}
	 & MAE & rMAE & CC \\ \hline
	\textbf{convGRU-CNN} & \boldmath $1.76 \pm 1.5$ & \boldmath $0.0213 \pm 0.0180$ & \boldmath $0.9996$ \\
	CNN-GRU & $2.06 \pm 3.4$ & $0.0249 \pm 0.0419$ & $0.9988$ \\
	2D CNN & $2.09 \pm 3.6$ & $0.0252 \pm 0.0440$ & $0.9987$ \\
	1D CNN & $3.24 \pm 4.0$ & $0.0392 \pm 0.0488$ & $0.9980$ \\
	GRU & $3.22 \pm 4.1$ & $ 0.0389 \pm 0.0490$ & $0.9980$ \\\hline \\
	\end{tabular}}
	\caption{Comparison of different models. The MAE  in $\si{\milli\newton}$ and rMAE (with standard deviation) and the CC are shown. Needle 1 was used for this experiment.}
	\label{tab:models}
\end{table}
Last, we show results for a needle insertion experiment for two different scenarios in Figure~\ref{fig:tissue}. One experiment was conducted with the shielding tube and one without. With the tube, predicted values closely match the measured values. Without, there are large deviations as the sensor also measures friction forces. 

\section{Discussion}

We introduce a novel technique for needle tip force estimation using an OCT fiber that images the deformation of an epoxy layer. As OCT has been used for needle-based tissue analysis \cite{Otte.2014}, it may become more widely available in clinical settings. Our method comes with the typical advantages of optical methods, such as MRI-compatibility and bio-compatibility, while also being flexible and easy to manufacture. This is highlighted by the results for three different needles with epoxy layers of different stiffness.  All needles show similar relative calibration errors with a CC in the range of $0.9996$ to $0.9989$, indicating that our method generalizes well for different epoxy resins. This allows for easy adaptation of our method to different scenarios with different requirements for force sensitivity and range.

Moreover, we propose a novel method for processing the spatio-temporal OCT data. 
Previously, time series of A-scans have been processed using recurrent architectures \cite{Otte.2014}. The approach shares parameters over time, however, it lacks effective spatial exploitation with parameter sharing over space through convolutions. As our convGRU model takes care of efficient processing of both temporal and spatial information, it outperforms the pure temporal GRU and pure spatial 1D CNN with an MAE of $1.76 \pm 1.5$ compared to an MAE of $3.24 \pm 4.0$ and $3.22 \pm 4.1$, respectively. Furthermore, we adopted a CNN-GRU and 2D CNN model from non-medical domains for comparison of spatio-temporal processing architectures \cite{donahue.2015long,xingjian2015convolutional}. Compared to the other models, the convGRU units in our model enable temporal processing first while keeping the data structure intact for subsequent CNN processing. This leads to superior performance for the problem at hand. 

Lastly, we tested our needle in an ex-vivo experiment with a human prostate. Several other needle tip force estimation methods have been proposed, however, they often lack validation in tissue experiments \cite{mo.2017capability}. One of the reasons for this is the difficulty to measure pure tip forces inside tissue as large friction forces will also be captured by external force sensors \cite{kataoka.2002measurement}. Therefore, we use a shielding tube that decouples friction forces from the needle. Although the decoupling is not perfect due to deformations, we can show that our method accurately captures events such as ruptures. Without the mechanism, in-tissue evaluation is not possible as frictional forces overlap with tip forces. The results indicate that our method is usable for actual force estimation in soft tissue. 

\section{Conclusion}

We propose a novel method for needle tip force estimation. Our approach uses an OCT fiber imaging the deformation of an epoxy layer to infer the force that acted on the needle tip. The concept is easy to realize and allows flexibility by using different materials for different force sensitivity and maximum range requirements. In order to process the spatio-temporal OCT data we propose a novel convGRU-CNN architecture. 
For our problem, the method outperforms prior approaches for similar problems and also methods from other domains. Experimental results for force estimation in human prostate tissue underline the method's potential for practical application.

\small \textbf{Acknowledgments.} This work was partially supported by DFG grants SCHL 1844/2-1 and SCHL 1844/2-2.

\bibliographystyle{spmpsci}      
\bibliography{egbib} 

\end{document}